# Transformationally Identical and Invariant Convolutional Neural Networks by Combining Symmetric Operations or Input Vectors

*ShihChung B. Lo, Ph.D.[1,2,3], Matthew T. Freedman, M.D.[2,3], and Seong K. Mun, Ph.D.[2]

[1] Radiology Department, Georgetown University Medical Center, Washington DC 20007
[2] Arlington Innovation Center: Health Research, Virginia Tech, Arlington VA 22203
[3] Oncology Department, Georgetown University Medical Center, Washington DC 20007
*e-mail: <dcben0@gmail.com> or <benlo@vt.edu>

## Abstract

Transformationally invariant processors constructed by transformed input vectors or operators have been suggested and applied to many applications. In this study, transformationally identical processing based on combining results of all sub-processes with corresponding transformations at one of the processing steps or at the beginning step were found to be equivalent for a given condition. This property can be applied to most convolutional neural network (CNN) systems. Specifically, a transformationally identical CNN can be constructed by arranging internally symmetric operations in parallel with the same transformation family that includes a flatten layer with weights sharing among their corresponding transformation elements. Other transformationally identical CNNs can be constructed by averaging transformed input vectors of the family at the input layer followed by an ordinary CNN process or by a set of symmetric operations. Interestingly, we found that both types of transformationally identical CNN systems are mathematically equivalent by either applying an averaging operation to corresponding elements of all sub-channels before the activation function or without using a non-linear activation function.

## 1. Introduction

Historically speaking, the use of transformed input vectors for data augmentation in the CNN was initiated by Lo and his co-workers (1993; 1995(a); 1995(b)). It is mathematically equivalent to use reversed transformation kernels as far as the convolution process is concerned. It is also more technically convenient to process the rotated kernels inside the CNN than a data augmentation approach that usually performs the process as additional sample cases. Samples of applying rotated kernels inside a CNN system are (1) Dieleman et al. (2016) who used cyclic symmetry approach to construct a CNN. (2) Cohen et al. (2016) demonstrated that rotated kernels can be formed as a group to construct and train a CNN system with a greater generation performance. Other publications related to the construction of transformationlly invariant neural networks qualitatively were contributed by recent investigators (Sifre 2013, Liao 2013, Gens 2014, Macos 2016, Weiler 2018, Cohen 2018). However, no author has shown a CNN system that can produce quantitatively invariant (identical) result through transformation until recently. Lo et al (2018(a)) showed that several transformationally identical and invariant CNN systems can be constructed through symmetric operator sets of the same transformation family. As a part of this research direction related to the transformation process within the CNN, this study is intended to demonstrate that transformationally identical and invariant CNN can also be constructed by combining (i.e., averaging in this study) symmetric operations or input vectors. These methods are mathematically related but different from the CNN system constructed by symmetric operators of the same transformation family.

All these approaches designed to produce transformationally identical results in a CNN are based on a transformation of the input vector that does not involve interpolation (type-1 transformation). In the same time, transformations of the input vector involving interpolation (type-2 transformation) would be used to train the CNN to produce qualitatively invariant output results or included in a further process for more comprehensive TI system composition (Lo 2018(b)). Since each type-2 transformation would have its associated type-1 transformation counterparts for training which is significantly different from conventional data augmentation approach. Each transformation of the input vector works as a group in the transformationally identical system. However, each transformed version of the input vector works on its own in the conventional method.

## 2. Thermotical Development and Methods
### 2.1. Scanning operations with and without symmetric element operators

The mathematical characteristics of transformationally identical process is broadly observed in various scan operations and functions. A well-rounded TI operator theorem and its special case in TI-CNN with the same family of symmetric kernels have been presented by Lo et al (2018(a)). Without using symmetric operators consistently, TI property may not be held with a series of scanning operation processes such as the CNN. However, a non-symmetrical operator can be considered as a composed operator with several symmetric ones. Hence, a TI property can be established if symmetric operations based on multiple sections of corresponding conceptually symmetrical operators arranged in parallel within the CNN.

### 2.2. Construction of parallel operations with non-symmetric operators of the same transformation family

Considering a vector ($V_i$) operated by an operation " ! " with a kernel "K" scanning through the entire $V_i$ space that results in another vector $V_{r1}$, as

$$V_{r1}(x',y',z') = V_i(x,y,z) ! K(u,v,s). \qquad ...(1)$$

where (u,v,s) denotes each element of K containing either a value or a function, but

$$K(u_t,v_t,s_t) \neq T\{ K(u_t,v_t,s_t) \} \qquad ...(2)$$

where T**{.}** is a transformation function of a vector.

By composing processes with rotated/reflected kernels as a group scanning over the same operand, they would produce a group of resultant matrices. This operation, functioning as a group, would produce quantitatively invariant result with an input vector of the same transformation family. The method is not much different from the conventional input data augmentation method with a type-1 transformation, except the grouped approach shares each operator with other corresponding transformations. It provides a platform for all input vector versions of rotation/reflection to be processed symmetrically and in parallel. Therefore, quantitative invariance property of the output through a transformation of input or operator is reachable. Common examples of using this rotation operation are (a) the use of 90° rotation as an increment with respect to each center axis (z=0, x=0, or y=0) as a family and (b) the use of 90° rotation increment and their reflections as a family (i.e., Dih4 symmetry family). In 2D, other rotation symmetric families are: 180-degree rotation, top-down reflection, left-right reflection, upper-left and lower-right reflection, and upper-right and lower-left reflection. There are many more symmetric families in 3D geometry as discussed in our recent paper (Lo 2018).

At the end of scanning operation (e.g., convolution), signals would then be passed to the non-convolution section. If the same output result is desired within the same transformation family, it is expected that signals from this point would be processed by:

   A.  Combining scan operation results from all corresponding rotated/reflected operations.

B. Equally treating each operation with a rotated/reflected operand or operator as a part of group processes (i.e., a pre-defined transformation family) until a later combining process that would result in the same final output when any input vector version of the same transformation family is entered in the input layer.

2.3. Construction of TI-CNN-2 with parallel kernel operations of the same transformation family

The discussion stated in previous section can be applied to construct a transfromationally identical CNN (TI-CNN-2) using regular convolution kernels and their rotations. This type of transfromationally identical CNN is different from the transfromationally identical CNN (TI-CNN-1) that is constructed by transfromationally invariant kernels (i.e., $K_h = T\{K_h\}$, for every hidden convolution layer "h"). In order to construct a rotationally identical CNN system, each rotated or flipped kernel is used to convolve the input vector $V_i(x,y,z)$ and would go through a typical CNN processing as an independent sub-channel. The intermediate result of each sub-channel can be expressed as:

$$V_{r1/a/r/f}(x',y',z') = V_i(x,y,z) * T_{a/r/f}\{K(u,v,s)\} \qquad ...(3)$$

where " * " denotes as a convolution operation. $T\{.\}$ is a rotation transformation function. "a" denotes major axis (z=0, y=0, and x=0), "r" represents rotation angle (0°, 90°, 180°, 270° with respect to one of major axie) and "f" represents with or without flipping (f: 0 or 1 with respect to the major axis). The kernel "K" used in all sub-channels are not the same but is the rotated versions corresponding to a/r/f. Its inversed rotation can turn it back to the original kernel

$$IT_{a/-r/cf}\{T_{a/r/ff}\{K(u,v,s)\}\} = K(u,v,s) \qquad ...(4)$$

where $IT_r\{.\}$ is the inverse transformation of $T_r\{.\}$. By replacing (4) in (3), we have

$$V_i * T_r\{K_{h1}\} = T_r\{IT_r\{V_i\}\} * T_r\{K_{h1}\} \qquad ,,,(5)$$

Since transformation and convolution operation are commutative, Eq. (5) can be extended

$$V_i * T_r\{K_{h1}\} = T_r\{IT_r\{V_i\}\} * T_r\{K_{h1}\} = T_r\{IT_r\{V_i\} * K_{h1}\} \qquad ...(6)$$

If the system only consists of one convolution process channel with kernels of the same transformation family, the resultant value received at the corresponding node (node-r) contributed from the "r" rotation sub-channel at the first flatter layer is

$$V_{node-r} = V_i * T_r\{K_{h1}\}... * T_r\{K_{hn}\} \cdot T_r\{K_f\} = T_r\{IT_r\{V_i\} * T_r\{K_{h1}\}... * T_r\{K_{hn}\} \cdot T_r\{K_f\}$$

$$= T_r\{IT_r\{V_i\} * K_{h1}... * K_{hn} \cdot K_f\} = IT_r\{V_i\} * K_{h1}...* K_{hn} \cdot K_f \qquad ...(7)$$

The 5th (last) set of Eq. (7) is due to that a transformation acts as an identity operator for a single value. The combined value from all sub-channel of transformed kernels

$$V_{node} = \sum_{r=1}^{M} V_i * T_r\{K_{h1}\} ... * T_r\{K_{hn}\} \cdot T_r\{K_f\} = \sum_{r=1}^{M} T_r\{IT_r\{V_i\}\} * T_r\{K_{h1}\} ... * T_r\{K_{hn}\} \cdot T_r\{K_f\}$$

$$= \sum_{r=1}^{M} IT_r\{V_i\} * K_{h1} ... * K_{hn} \cdot K_f = \sum_{r=1}^{M} IT_r\{V_i\} * T_j\{K_{h1}\} ... * T_j\{K_{hn}\} \cdot T_j\{K_f\} \qquad ...(8)$$

where "M" is 8 and 24 for 2D and 3D Dih4 symmetry families, respectively. j=1, 2, …, or M. $T_j\{.\}$ denotes any rotation version of the Dih4 family. For the right-angle rotation family, "M" is 4 and 12 for 2D and 3D CNNs, respectively.

Since convolution processes are independent from the operation of combining all transformations, Eq. (8) can also be rewritten as

$$V_{node} = \sum_{r=1}^{M} Vi * T_r\{K_{h1}\} ... * T_r\{K_{hn}\} \cdot T_r\{K_f\}$$
$$= (\sum_{r=1}^{M} IT_r\{Vi\}) * K_{h1} ... * K_{hn} \cdot K_f = (\sum_{r=1}^{M} IT_r\{Vi\}) * T_j\{K_{h1}\} ... * T_j\{K_{hn}\} \cdot T_j\{K_f\} \quad ...(9)$$

If every convolution process followed by a non-linear activation and/or pooling (denotes as KC), the results of these two processes would not be the same any more.

$$\overline{V'}_{node} = (\sum_{r=1}^{M} Vi * T_r\{KC_{h1}\} ... * T_r\{KC_{hn}\} \cdot T_r\{K_f\})/M \quad ...(10)$$

for averaging results at the end of convolution and activation processes

$$\overline{V''}_{node} = ((\sum_{r=1}^{M} IT_r\{Vi\})/M) * KC_{h1} ... * KC_{hn} \cdot K_f$$
$$= ((\sum_{r=1}^{M} IT_r\{Vi\})/M) * T_j\{KC_{h1}\} ... * T_j\{KC_{hn}\} \cdot T_j\{K_f\} \quad ...(11)$$

for averaging transformed input vectors at the input layer.

After this point, signals are processed by same sets of network weights. Based on Eqs. (10) and (11), we can further conclude that

$$CNN_{(\sum_{r=1}^{M} T_r\{K\})/M}[Vi] = CNN_{(\sum_{r=1}^{M} T_r\{K\})/M}[T_j\{Vi\}] = V'_o \quad ...(12)$$

for averaging results at the end of convolution and activation processes with further network processes up to the output layer.

$$CNN_{Tj\{K\}}[(\sum_{r=1}^{M} IT_r\{Vi\})/M] = CNN_{Tj\{K\}}[(\sum_{r=1}^{M} T_r\{Vi\})/M] = V''_o \quad ...(13)$$

for averaging transformed input vectors at the input layer followed by an ordinary CNN process. For averaging results at one set of convolution with activation processes followed by an ordinary CNN process, a version of transformationally identical TI-CNN-2 can also be established.

Again, $T_j[.]$ denotes any "j" set of the transformation and j=1,2...or M. Therefore, there are 8 and 24 versions of $CNN_{Tj\{K\}}[(\sum_{r=1}^{M} IT_r\{Vi\})/M]$ can be constructed with Dih4 symmetry kernels in 2D and 3D geometries, respectively. This theoretical derivation demonstrates that each of these two CNN architectures with the same sets of kernels would produce an identical output result for a transformed input vector of the same transformation symmetry family.

1) Each convolution process is spitted into multiple processes with kernels of the same transformation family (rotated/reflected versions of the kernel) and equally combined at the end or at middle of convolution pipelines. Fig. 1 shows the schematic diagram of the first CNN structure (TI-CNN-2-1).

2) The input vector is processed by each of multiple transformation versions of the same family (rotated/reflected versions of the input vector) and equally combined (averaging) at the input layer. The combined input vector is then processed by an ordinary CNN processing. Fig. 2 shows the schematic diagram of the second CNN structure (TI-CNN-2-2).

The 2nd CNN architecture is basically the same as an ordinary CNN except averaging all transformation versions of the input vector before getting into CNN processes. Furthermore Eq. (13) holds for any arbitrary input vector version of the transformation family.

$$CNN_{Tj\{K\}}[(\sum_{r=1}^{M} IT_r\{T_b\{Vi\}\})/M] = CNN_{Tj\{K\}}[(\sum_{r=1}^{M} T_r\{T_b\{Vi\}\})/M] = V''_o \quad ...(14)$$

where $T_a\{.\}$ and $T_b\{.\}$ denote any "a" and "b" version of the transformation and "a" or "b" =1,2...or

M. The TI-CNN-2-1 system (averaging results at the first flatten layer) and the TI-CNN-2-2 system (averaging inputs at the input layer) would produce identical result if no non-linear activation is used. Because averaging operation and non-linear activation function are not commutative, corresponding intermediated and final results would not be the same. But they can be made equivalent by averaging intermediated results prior to each activation function. In other words, by altering convolution/activation operation to convolution/averaging/activation operation at each element in this special type of TI-CNN-2-1 system (or TI-CNN-2-1.2), the corresponding intermediated and final results would be the same as that of the TI-CNN-2-2 system with the same corresponding convolution processes after the first convolution layer.

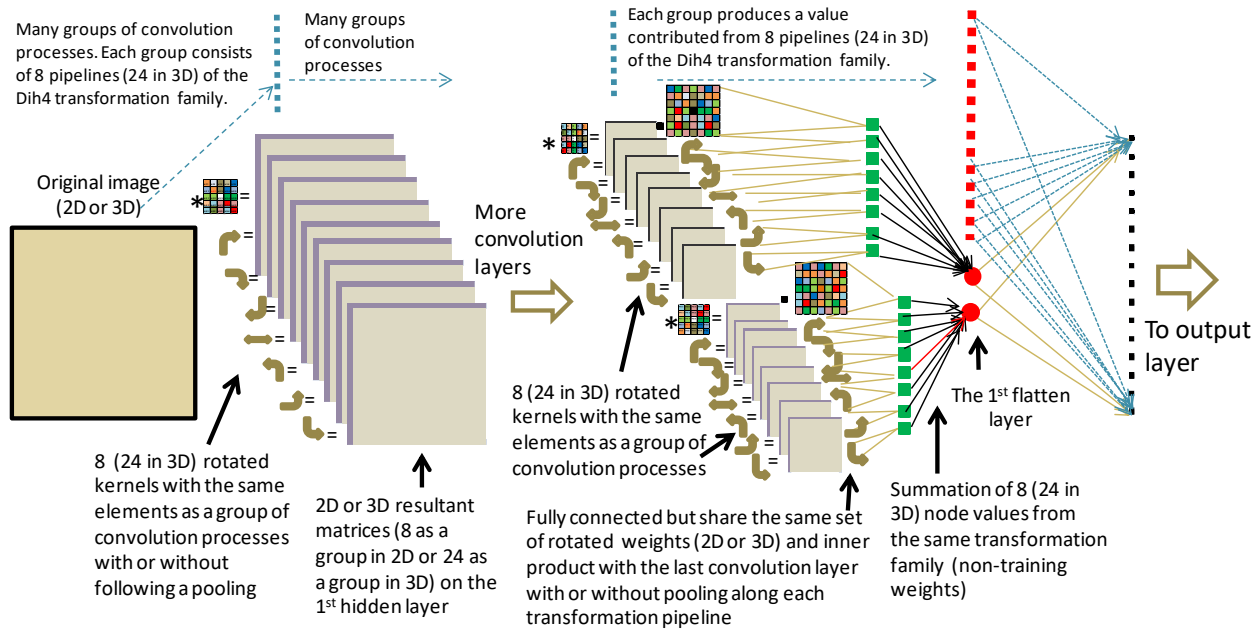

Figure 1. Averaging results from all sub-channels at one set of convolution with activation layers or at the end of feature terminal layers (i.e., the first flatten layer). Each sub-channel is processed with a set of same transformation kernels. ⌐ ⌐ ⌐ ⇌ ⌐ ⌐ ⌐ represent rotations with 90º, 180º, 270º, reflection, 90º & reflection, 180º & reflection, and 270º & reflection, respectively.

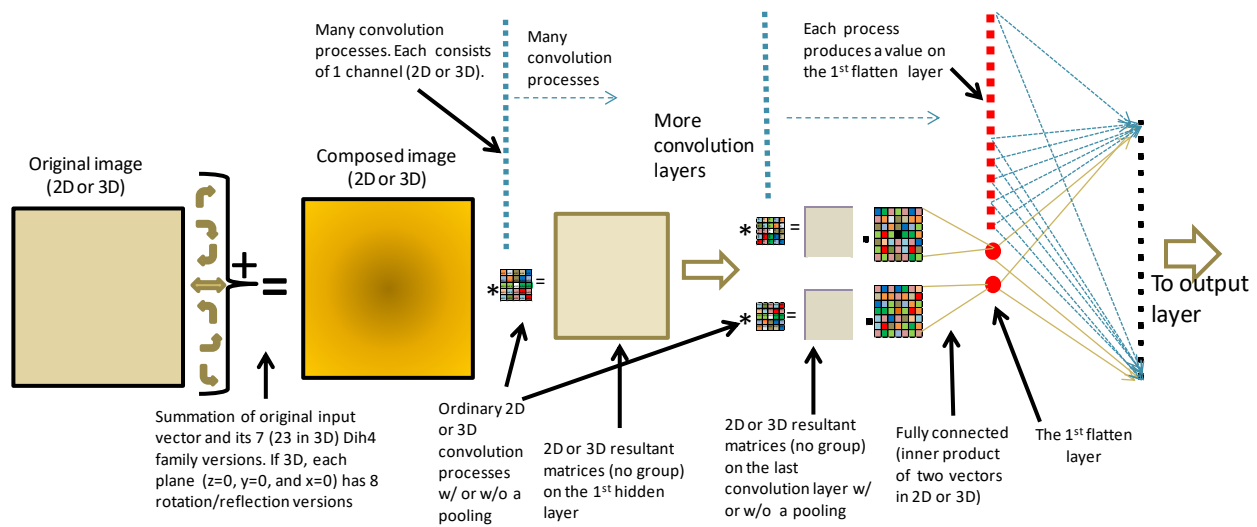

Figure 2. Averaging all transformed input vector versions at the input layer first and then processed the composed input vector through an ordinary CNN process.

2.4. The characteristics of intermediate resultant matrices at each hidden layer and their implications of the error-backpropagation training

By rotating or flipping the input vector or all kernel operators to form a CNN with multiple sub-channels, each sub-channel would produce a resultant matrix with or without using an activation function and other processes at each convolution layer. If we average these resultant matrices (8 in 2D, 24 in 3D) of the same group into one matrix in TI-CNN-2-1, the combined intermediate resultant matrix is different from the corresponding intermediate resultant matrix on the corresponding channel of the hidden layer in TI-CNN-2-2. But the TI-CNN-2-1.2 system would produce the same results as a TI-CNN-2-2 system possessing the same set of convolution processes after the first convolution layer.

For the application of using rotated kernels over each convolution in a CNN, we also need to consider about training of the combined convolution process with the error-backpropagation. Should the averaging or summation operation be used for combining processes, the error-backpropagation method would still be able to serve as a training process as experienced in an ordinary CNN using multiple channels internally. But corresponding weight update processes in the TI-CNN-2-1 and the TI-CNN-2-2 do not seem to have any sharable operation due to non-commutative between activation and averaging operations. Particularly, intermediate matrices on any of the corresponding hidden layers are different in both systems. However, a high-order regression process is possible to link each of these non-commutative averaging/activation operations which is beyond the scope of this study.

For the TI-CNN-2-2 system, the error-backpropagation training is the same as an ordinary CNN. For the TI-CNN-2-1 system, each update step though error-backpropagation for training a kernel used in a sub-channel should be combined by an averaging operation. Hence, the weights would be updated differently between the TI-CNN-2-1 and the TI-CNN-2-2. Therefore, it is difficult to predict that the final sets of kernel weights trained by these two CNN structures may result in the same or very close.

2.5. Characteristics of TI-CNN-2 systems

Several characteristics of the TI-CNN-2 may or may not be shared with the TI-CNN-1 and are worth mentioning that:

1) TI-CNN-2 systems is based on averaging symmetric processes or transformed input vectors. Unlike the TI-CNN-1, no constraint is necessary to impose on any kernel and weight besides error-backpropagation training. It is expected that these TI convolution kernels may be able to extract features associate with the TI likely representing a symmetric property in the input and may reflect this TI on the output.

2) No need to use inclusive transformation versions of input vector T**{** Vi **}** as a part of data augmentation for the training of a TI-CNN system.

3) Unlike the TI-CNN-1 (use of symmetric operators), the total effective number of elements in a kernel is not changed in the TI-CNN-2-1 or the TI-CNN-2-2 as compared to its ordinary CNN counterpart.

4) During the training, the TI-CNN systems (TI-CNN-1, TI-CNN-2-1, and TI-CNN-2-2) intrinsically takes multiple input versions with each type-2 transformation input. For example, by rotating the Vi with $d°$ and entering it onto the input layer, the training for a 2D CNN with Dih4 kernels would automatically gain a total of 8 versions including $d°$, $(d+90)°$, $(d+180)°$, $(d+270)°$, and their reflections. Again, the type-2 transformation of the input vector can be used to train the TI-CNN-2-1 and the TI-CNN-2-2 to achieve qualitatively invariant output results. It can also be used for further development of a comprehensive TI-CNN structure. When both type-1 and type-2 transformations are used in each of the proposed TI-CNN systems, it is expected to produce a greater generalization performance.

## 3. Experiments and Initial Results of 2D TI-CNN-2s

We used random number generator to perform the study and to valid the proposed transformationally identical systems with a wide range of parameter settings for a variety of CNN structures composing of convolution kernels with and without involving stride, pooling, 1D convolution across the channels in a CNN layer. Specifically, we randomly generated 2D matrix $V_i(x,y)$ and 2D kernels $K_{l,c}(u,v)$ where "l" denotes layer number and "c" denotes channel number in layer "l", $T_{tr}\{\cdot\}$ transformation of a matrix with either 90°, 180°, 270°, or 360° rotation and their corresponding reflection counterparts. The size of $V_i$ ranged from 10 to 1000 in 1D and kernel size ranged from 3 to 30 in 1D, number of convolution layers ranged from 1 to 10. The size of channel matrices in any hidden convolution layer are evenly divisible by the stride length and its immediate pooling size.

The result indicated that the difference between two output vectors: CNN-2-1[ $V_i$ ] - CNN-2-1[$T_{tr}\{ V_i \}$] as well as CNN-2-2[ $V_i$ ] - CNN-2-2[$T_{tr}\{ V_i \}$] vanished (i.e., $\vec{0}$) for all tests with type-1 rotations and their reflections as well as all sizes of $K_{l,c}(u,v)$ that contain different sets of convolution kernels and weights.

When cross-comparing the TI-CNN-2-1.2 and the TI-CNN-2-2 structures, all results showed that CNN-2-1.2[ $V_i$ ] - CNN-2-2[$V_i$] = $\vec{0}$ as well. The same results were also obtained with all versions input vector with or without rotation: CNN-2-1.2[ $T_a\{V_i\}$ ] - CNN-2-2[ $T_b\{V_i\}$ ] = $\vec{0}$. where $T_a[.]$ and $T_b[.]$ represent two arbitrary transformations of the Dih4 family. These results clearly confirm the TI-CNN-2 theorem development described in section 2. They also indicated that the proposed CNN systems are transformationally identical when an input vector of the same transformation family is entered in the input layer.

## 4. Conclusions and Discussion

Both theoretical derivations and experimental results demonstrated that proposed two CNN structures produce identical output vector providing that they share the same sets of convolution kernels and connected weights. Hence, we can clearly claim that they are TI functions and named them TI-CNN-2-1 and TI-CNN-2-2, respectively. A list of the TI families for 2D and 3D geometries is shown in our previous paper [Lo 2018].

The implementation of TI-CNN-2-2 is no difference from an ordinary CNN except averaging all input vectors of the same transformation family at the input layer before processing it with a series of the CNN processes in its forward propagation and its back propagation for the training as shown in Fig 2. The implementation of TI-CNN-2-1 would require splitting each convolution process into 8 and 24 sub-processes for the Dih4 symmetry family in 2D and 3D systems, respectively. The corresponding sub-processes would be merged in the final feature termination layer (i.e., the first flatten layer). The backpropagation for the TI-CNN-2-1 would require an extra operation in each convolution step that updated weights from each sub-channel of a group convolution processes should be shared with others in the transformation group. The implementation of TI-CNN-2-1.2 would require merging convolution results of sub-channels and repeat the splitting and merging processes in subsequent convolution layers.

We were also curious to compare TI-CNN-1 and TI-CNN-2 and found that TI-CNN-1 produced different results from the TI-CNN-2-1 and the TI-CNN-2-2. This implies that TI-CNN-1 and TI-CNN-2 can be composed together to form another TI-CNN system (i.e., TI-CNN-C) using the same family of transformation kernels. Finding relations between TI-CNN-1, TI-CNN-2-1, TI-CNN-2-2, and TI-CNN-C and their performances in various applications are called for further investigation. Again, all TI-CNNs do not need to recalculate when one of input vector version of the same transformation family has been used. However, type-2 transformations can still be used for training as a part of data augmentation; and each input vector with type-2 transformation would inherently include its members of transformation family in the training. They can also be included in a more comprehensive TI system composition.

4.1. Potential Issues of using TI-CNN-2 systems (both TI-CNN-2-1 and TI-CNN-2-2)

The input vector entered in the input layer, though it can be any version of the transformation family, in effect would produce the same output result as the input vector composed of all or a part of transformed input vector versions. In other words, the use of TI-CNN-2-1 or TI-CNN-2-2 would not be able to distinguish the input vector and input vectors composed by its associated transformation family members. Because these input vector would create the same composed input vector patterns with the formulation of ( $\sum_{r=1}^{M} T_r\{T_b\{Vi\}\})/M$ indicated in Eq. (13) before a series of convolution processes in an ordinary CNN. Though TI-CNN-2-1 is not started with composing input vectors of the same transformation family, each group (8 and 24 sub-channels in 2D and 3D Dih4 families, respectively) of convolution process would be merged at the first flatten layer. The merger would in effect make that all associated versions of input vectors and their compositions result in the same vector pattern, though differ in intensity, before the CNN processing. They may produce the same output, if a normalization function (such as softmax) is used at one point of the CNN procedure. This is particularly true for the two-tone image pattern recognition. If the original image does not possess any symmetric component, a total of 2,160 patterns would produce the same output result using the TI-CNN-2-1 or the TI-CNN-2-2 system based on the recognition requirement of 2-D Dih4 symmetry. Fig. 3 shows an input vector (an arrow) and some of its associated versions composed by a part of the same transformation family.

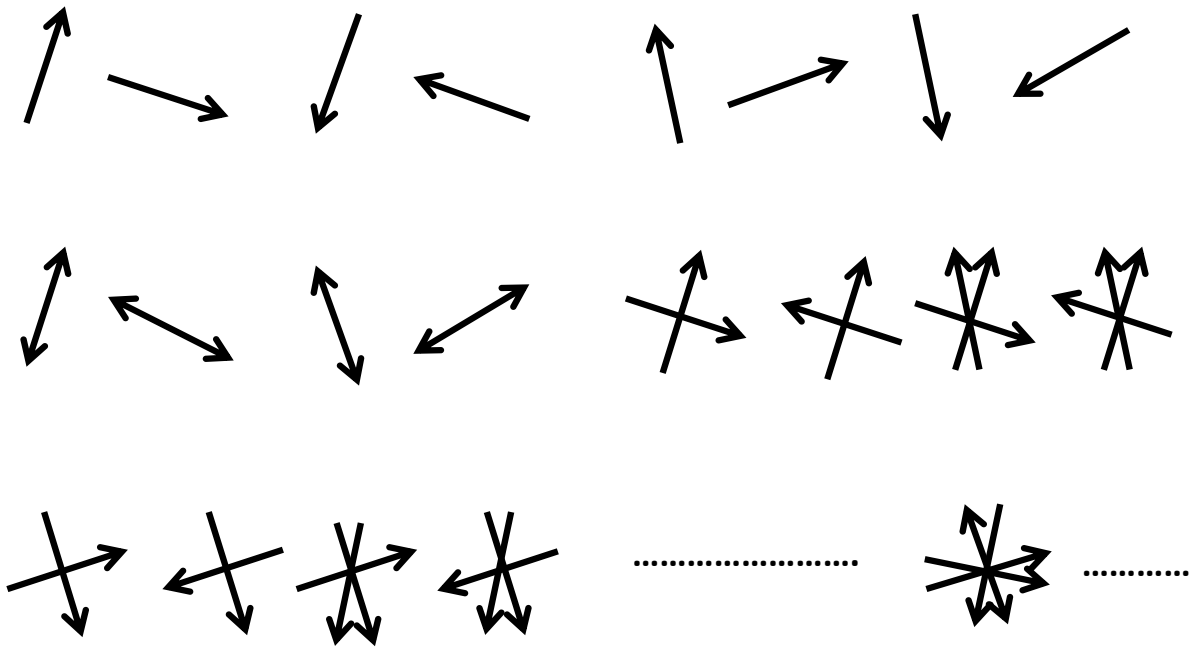

Figure 3. Any input vector (say the one at the left upper corner), its transformation family members (the first row) and their compositions (totally 2,160 input vectors) would generate the same output result when using TI-CNN-2-1 or TI-CNN-2-2.

## References

Cohen, T.S., and Welling, M., "Group equivariant convolutional networks." In Proceedings of The 33rd International Conference on Machine Learning (ICML), volume 48, pages 2990–2999, 2016.

Cohen, T.S., Geiger M., Koehler, J., and Welling, M., "Spherical CNNs." in ICLR 2018.

Dieleman, S., De Fauw, J., and Kavukcuoglu, K., "Exploiting Cyclic Symmetry in Convolutional Neural Networks." In International Conference on Machine Learning (ICML), 2016.

Gens, R. and Domingos, P., "Deep Symmetry Networks." In Advances in Neural Information Processing Systems (NIPS), 2014.

Liao, Q., Leibo, JZ, and Poggio, T., "Learning invariant representations and applications to face verification. In Advances in Neural Information Processing Systems, pp. 3057–3065, 2013.

Lo, S.B., Lin J.S., Freedman, M.T., and Mun, S.K., "Computer-Assisted Diagnosis of Lung Nodule Detection using Artificial Convolution Neural Network", SPIE Proc. Med. Imag. VII, vol. 1898, 1993, pp. 859-869.

Lo, S.B., Lou, S.L., Lin, J.S., Freedman, M.T., Chien, M.V., and Mun, S.K., "Artificial Convolution Neural Network Techniques and Applications to Lung Nodule Detection," IEEE Trans. Med. Imag., 1995(a), vol. 14, No. 4, pp. 711-718.

Lo, S.B., Chan, H.P., Lin, J.S., Li, H., Freedman, M.T., and Mun, S.K., Artificial Convolution Neural Network for Medical Image Pattern Recognition, Neural Networks, 1995(b), Vol. 8, No. 7/8, pp. 1201-1214.

Lo, S.B., Freedman, M.T., Mun, S.K., Gu, S., "Transformationally Identical and Invariant Convolutional Neural Networks through Symmetric Operators," arXiv, June 2018 (a), https://arxiv.org/ftp/arxiv/papers/1806/1806.03636.pdf

Lo, S.B., Freedman, M.T., Mun, S.K., Chen, H.P., "Geared Rotationally Identical and Invariant Convolutional Neural Network Systems," arXiv, August 2018 (b), https://arxiv.org/ftp/arxiv/papers/1808/1808.01280.pdf

Marcos, D., Volpi, M., and Tuia, D., "Learning rotation invariant convolutional filters for texture classification, 2016, https://arxiv.org/pdf/1604.06720.pdf

Sifre, L., and Mallat, S., "Rotation, scaling ´ and deformation invariant scattering for texture discrimination". In Computer Vision and Pattern Recognition (CVPR), 2013 IEEE Conference on, pp. 1233–1240. IEEE, 2013.

Weiler, M., Hamprecht, F.A., and Storath, M., "Learning steerable filters for rotation equivariance CNNs." 2018. https://arxiv.org/pdf/1711.07289.pdf

Worrall, D.E., Garbin, S.J., Turmukhambetov D., and Brostow G.J., "Harmonic networks: Deep translation and rotation equivariance." In CVPR, 2017.